\documentclass{article}

\usepackage[dblblindworkshop,final]{neurips_2026}

\usepackage{subcaption}
\usepackage[utf8]{inputenc} 
\usepackage[T1]{fontenc}    
\usepackage[hidelinks]{hyperref}       
\usepackage{url}            
\usepackage{booktabs}       
\usepackage{amsfonts}       
\usepackage{nicefrac}       
\usepackage{microtype}      
\usepackage{xcolor}         
\usepackage{amsmath}        
\usepackage{multirow}      
\usepackage{graphicx}
\usepackage{array}
\usepackage{longtable}
\hypersetup{pdfauthor={Nazim-E-Alam, Tarek Rahman, Md Kishor Morol, Jannatun Noor},pdftitle={AAMBERS-UAV: Acquisition-Aware Multimodal Backbone Evaluation and Ranking for UAV Weedy Rice Segmentation}}
\setcitestyle{square,numbers,comma}
\title{AAMBERS-UAV: \underline{A}cquisition-\underline{A}ware
\underline{M}ultimodal \underline{B}ackbone \underline{E}valuation and
\underline{R}anking for UAV Weedy Rice \underline{S}egmentation}

\workshoptitle{The 7th Muslims in ML (MusIML) Workshop at NeurIPS 2026}

\author{%
Nazim-E-Alam$^{2}$ \qquad Tarek Rahman$^{1,3}$ \\
Md Kishor Morol$^{3}$ \qquad Jannatun Noor$^{1}$ \\[5pt]
$^1$United International University \\[2pt]
$^2$American International University Bangladesh \\[2pt]
$^3$ELITE Research Lab, New York, USA
}
\begin{document}

\maketitle

\begin{abstract}
UAV image collections contain spatially and temporally related frames, yet semantic-segmentation benchmarks commonly split them at image level. Such splitting can place samples from one acquisition in both model development and testing, obscuring transfer to a genuinely new survey. Using the 734-sample WeedyRice-RGBMS-DB, we fix a 124-image target-acquisition test set and compare two protocols with identical train, validation, and test counts: target-held-out, which excludes the target acquisition from development, and target-exposed, which admits its remaining images. SegFormer-B0 is evaluated with RGB, four-band multispectral (MS), and seven-channel RGB+MS input over two fixed-split seeds. RGB is strongest under complete acquisition holdout ($0.7317\pm0.0201$ IoU), whereas RGB+MS becomes strongest after target exposure ($0.7822\pm0.0269$). A fixed-split U-Net/ResNet18 replication confirms positive exposure gains for all three inputs, but retains RGB as the best modality under both protocols. Acquisition exposure therefore increases measured performance across both evaluated backbones, while its effect on modality ranking is architecture-dependent. A supplied-split audit reveals strong near-sequential dependence, and corruption tests show that early fusion is substantially more sensitive to RGB--MS displacement than to moderate radiometric scaling. These results support acquisition-aware same-test evaluation as a necessary complement to ordinary image-level splitting in multimodal UAV benchmarks.
\end{abstract}
\section{Introduction}
Weedy rice is a persistent constraint in paddy production because it competes
with cultivated rice for light, water, and nutrients while remaining difficult
to distinguish during overlapping growth stages. Manual field scouting is
labour-intensive and often too slow for site-specific intervention. UAV sensing
provides centimetre-scale field coverage, flexible revisit times, and imagery
suitable for both mapping and prescription generation. Multispectral cameras
further add Red Edge and near-infrared information beyond conventional RGB,
motivating studies based on vegetation indices, RGB--MS fusion, and semantic
segmentation~\cite{stroppiana2018early,barrero2018rgb,lan2021realtime,yu2022weed}.
Dense crop--weed perception has also advanced through multispectral
convolutional networks and real-time agricultural vision
systems~\cite{sa2018weednet,milioto2018realtime,fraccaro2022deep}.

WeedyRice-RGBMS-DB provides an unusually useful benchmark: 734 pixel-aligned
UAV samples containing RGB, Green, Red, Red Edge, NIR, and a binary weedy-rice
mask~\cite{nguyen2025dataset}. It complements an earlier RGB-focused
comparison~\cite{nguyen2025detection} and recent few-shot work linking
segmentation to herbicide prescription maps~\cite{li2026fewshot}. However,
UAV images are not independent observations. Nearby frames from one flight can
share field content, illumination, crop stage, altitude, sensing geometry, and
sensor calibration. If such frames are separated only at image level,
development and test partitions can contain close representatives of the same
acquisition. The resulting score may therefore reflect interpolation within a
survey rather than transfer to a genuinely new flight or field condition.

This issue is particularly important for pixel-aligned multimodal fusion. A
model may exploit cross-modal relationships that are stable within one
acquisition but brittle when sensing geometry or field conditions change.
Structured-validation literature recommends blocking by temporal, spatial, or
hierarchical units when deployment requires prediction beyond observed
groups~\cite{roberts2017crossvalidation}. Remote-sensing studies likewise show
that spatial autocorrelation can inflate CNN evaluation~\cite{kattenborn2022spatial}
and that spatial validation can expose weak transfer hidden by conventional
splits~\cite{ploton2020spatial}.

These observations reveal an important evaluation gap: it remains unclear
whether the apparent benefit of multimodal input persists when the target
acquisition is completely unseen during development, and whether the resulting
modality ranking is stable across segmentation backbones. To address this gap,
we ask: \emph{Does target-acquisition exposure alter the ranking of RGB, MS,
and RGB+MS inputs, and is that ranking stable across backbones?} We isolate
this effect using a same-test design in which the target test set is held fixed
while only the availability of the remaining target-acquisition images during
model development is changed.

Our contributions are:
\begin{enumerate}
    \item an audit of temporal, frame-sequence, and geospatial dependence in
    the dataset's supplied split;
    
    \item a same-test evaluation protocol contrasting target-held-out and
    target-exposed development with equal sample counts;
    
    \item a two-seed SegFormer-B0 study with RGB, MS, and RGB+MS inputs,
    together with a fixed-split U-Net/ResNet18 backbone replication; and
    
    \item paired interaction, coverage, and sensor-corruption analyses that
    separate acquisition-exposure effects from architecture-dependent
    multimodal behaviour.
\end{enumerate}

\section{Related Work}
This section reviews prior work on UAV weed mapping, multimodal segmentation, and structured validation, establishing the context for evaluating acquisition-level dependence rather than relying only on image-level splits.
\paragraph{UAV weed mapping and multimodal segmentation.}
Early UAV weed-mapping studies combined multispectral orthomosaics and vegetation indices for rice management~\cite{stroppiana2018early,yu2022weed}. Barrero and Perdomo fused RGB texture with spectral information for Gramineae detection in rice fields~\cite{barrero2018rgb}; weedNet later demonstrated dense multispectral crop--weed classification from aerial imagery~\cite{sa2018weednet}. Semantic segmentation has also enabled real-time rice-weed recognition on embedded platforms~\cite{lan2021realtime}, while broader UAV weed-mapping work confirms the value of dense outputs for spatial management~\cite{fraccaro2022deep}. A recent review notes that the field still relies on relatively few benchmarks and offers limited evidence of transfer across environments~\cite{ahmad2026review}.

WeedyRice-RGBMS-DB is especially suitable for this question because its four MS bands are registered to the RGB image at $1280\times960$ resolution~\cite{nguyen2025dataset}. Earlier work on the collection focused on RGB-based detection and segmentation~\cite{nguyen2025detection}, while later work explored few-shot multimodal segmentation for prescription-map generation~\cite{li2026fewshot}. Our objective is complementary: rather than proposing a new fusion network, we isolate how acquisition exposure changes the measured benefit of modalities and whether that conclusion transfers between a Transformer and a conventional CNN.

\paragraph{Structured dependence in model validation.}
Ordinary random splitting implicitly treats observations as exchangeable. That assumption is inappropriate when samples are clustered in space, time, flight sequence, or acquisition units. Roberts et al.~\cite{roberts2017crossvalidation} recommend blocked validation whenever the deployment target lies outside observed structure. Kattenborn et al.~\cite{kattenborn2022spatial} quantified inflated CNN scores caused by spatially autocorrelated training and validation samples; Ploton et al.~\cite{ploton2020spatial} similarly found that spatial validation revealed much weaker ecological mapping transfer. Our design applies the same principle at acquisition level while keeping the target test set fixed.

\section{Experimental Design and Methodology}
\begin{figure*}[ht]
\centering
\includegraphics[width=.85\linewidth]{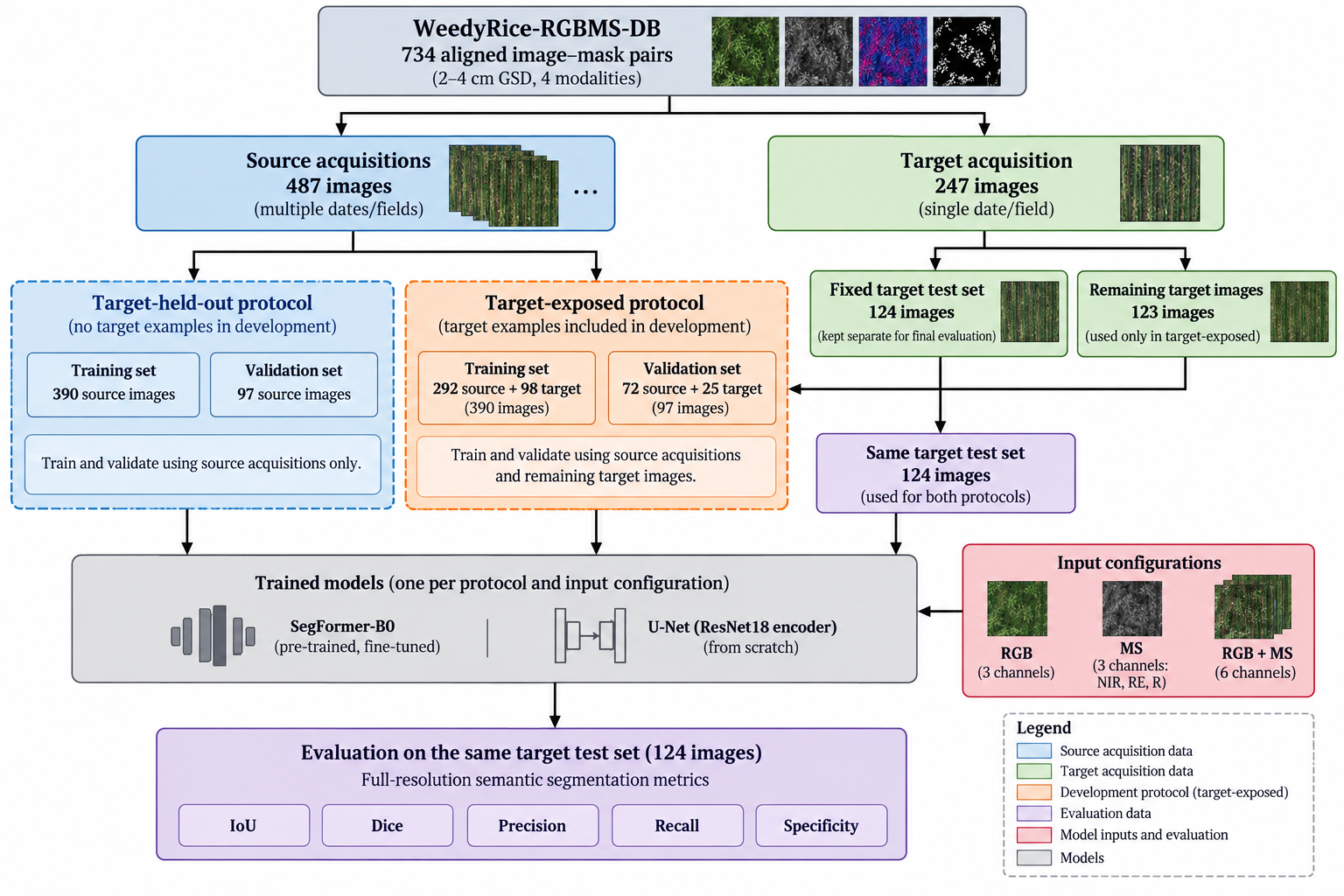}
\caption{Dataset split and acquisition-aware development protocols. The fixed
124-image target test set is identical for both protocols; the remaining
123 target-acquisition images are available only in target-exposed development.
Both protocols are evaluated on the same target test set across the three
input configurations and both evaluated backbones.}
\label{fig:protocol}
\end{figure*}
The methodology is organized around dataset auditing, controlled acquisition-aware splitting, multimodal model training, and full-resolution statistical evaluation.
\subsection{Dataset and Acquisition Audit}

The dataset contains 734 aligned image--mask pairs acquired during four UAV surveys in Vietnam's Mekong Delta across three cropping seasons~\cite{nguyen2025dataset}. Date, time, frame number, latitude, longitude, and altitude are recovered directly from the released filenames. The first three surveys contribute 487 source images; the survey collected on 15 January 2025 contributes 247 images and is treated as the target acquisition. Appendix~\ref{sec:dataset}, Table~\ref{tab:inventory} summarizes the acquisition inventory
and the supplied-split dependence statistics. This is an acquisition-domain shift, not a causal seasonal experiment. Mean annotated weed coverage rises from 0.181 in the earlier acquisitions to 0.585 in the target. Location, phenotype, illumination, infestation, and altitude therefore change together. A 64-bit difference hash is used only as a conservative image-similarity check; it does not establish pixel-identical duplication. Representative low-, medium-, and high-coverage samples are shown in Appendix~\ref{sec:dataset}, Figure~\ref{fig:examples}.

\subsection{Same-test Acquisition Protocols}
Figure~\ref{fig:protocol} summarizes the acquisition-aware development protocols.
A coverage- and altitude-stratified sample of 124 target images is fixed as the test set.
Both protocols contain exactly 390 training and 97 validation images.
\begin{itemize}
  \item \textbf{Target-held-out:} all 487 development images come from the three earlier acquisitions; no target image enters training or validation.
  \item \textbf{Target-exposed:} the fixed test set is unchanged. Development contains 292 source and 98 target training images, plus 72 source and 25 target validation images. Source images are subsampled so train/validation totals match the held-out protocol.
\end{itemize}
Splitting is performed at full-image level before any crop is generated. Thus, overlapping crops from one parent image cannot cross partitions. The same image assignments are used for SegFormer seeds 17 and 43 and for the U-Net replication.

\subsection{Inputs, Architectures, and Training}
Three input conditions are compared: RGB (3 channels), MS (Green, Red, Red Edge, NIR), and seven-channel early RGB+MS concatenation. The primary model is SegFormer-B0, whose MiT-B0 encoder and lightweight decoder provide a compact Transformer baseline~\cite{xie2021segformer}. To test backbone dependence, the complete six-cell matrix is repeated once with U-Net and a ResNet18 encoder, a conventional encoder--decoder reference~\cite{ronneberger2015unet}. For 4- and 7-channel inputs, the first encoder layer is adapted while retaining ImageNet initialization for the pretrained channels.

RGB uses ImageNet normalization. For MS channels, training-only 1st and 99th percentiles define clipping, after which scaling and standardization are fitted separately within each protocol. Training uses two on-the-fly $384\times384$ crops per image per epoch, batch size 8, horizontal and vertical flips, $90^\circ$ rotations, and a 0.55 probability of sampling a foreground-containing crop. The loss combines weighted binary cross-entropy and soft Dice in equal proportions; Dice provides an overlap-oriented objective under foreground imbalance~\cite{milletari2016vnet}. AdamW uses learning rate $2\times10^{-4}$ and weight decay $10^{-4}$~\cite{loshchilov2019adamw}. Runs use at most 45 epochs, ReduceLROnPlateau scheduling, patience 7, and mixed-precision training.

\subsection{Full-Resolution Evaluation and Statistical Analysis}
Each $1280\times960$ image is evaluated with overlapping $384\times384$ tiles, stride 320, and Hann-weighted logit blending. A single global threshold is chosen on validation images from a grid spanning 0.30--0.99 and then frozen for testing. IoU, Dice, precision, recall, and specificity are computed from aggregate full-resolution confusion counts; no crop-level test scores are reported.
The primary table reports mean $\pm$ standard deviation across two SegFormer model seeds. U-Net is a one-seed backbone replication. Paired image bootstrapping with 20,000 resamples quantifies protocol and modality differences on identical test images. These uncertainties are kept separate: seed standard deviation measures training variability, whereas image bootstrapping measures test-sample uncertainty conditional on one fitted model.

To directly test whether exposure changes fusion's relative value, we define
\begin{equation}
\Delta_{\mathrm{int}} = (I_{F,E}-I_{R,E})-(I_{F,H}-I_{R,H})
\label{eq:interaction}
\end{equation}
where $I$ is test IoU, $F$ denotes RGB+MS, $R$ denotes RGB, and $E/H$ denote exposed/held-out development. Positive $\Delta_{\mathrm{int}}$ means exposure increases fusion's advantage over RGB.
\begin{table*}[t]
\caption{Full-resolution SegFormer performance on the identical 124-image target test set (mean $\pm$ standard deviation over two fixed-split model seeds). Best IoU and Dice within each protocol are bold.}
\label{tab:segformer}
\centering
\scriptsize
\setlength{\tabcolsep}{3.6pt}
\begin{tabular}{llccccc}
\toprule
Protocol & Input & IoU & Dice & Precision & Recall & Specificity \\
\midrule
\multirow{3}{*}{Held-out}
& RGB & \textbf{.7317$\pm$.0201} & \textbf{.8450$\pm$.0134} & .8002$\pm$.0112 & .8950$\pm$.0162 & .6851$\pm$.0163 \\
& MS & .7184$\pm$.0059 & .8361$\pm$.0040 & .7913$\pm$.0027 & .8864$\pm$.0124 & .6704$\pm$.0101 \\
& RGB+MS & .7015$\pm$.0161 & .8245$\pm$.0111 & .7490$\pm$.0290 & \textbf{.9176$\pm$.0161} & .5648$\pm$.0745 \\
\midrule
\multirow{3}{*}{Exposed}
& RGB & .7666$\pm$.0328 & .8677$\pm$.0210 & .8259$\pm$.0351 & \textbf{.9144$\pm$.0037} & .7269$\pm$.0653 \\
& MS & .7348$\pm$.0142 & .8471$\pm$.0094 & .8287$\pm$.0045 & .8663$\pm$.0149 & .7476$\pm$.0036 \\
& RGB+MS & \textbf{.7822$\pm$.0269} & \textbf{.8777$\pm$.0170} & \textbf{.8479$\pm$.0216} & .9097$\pm$.0116 & \textbf{.7697$\pm$.0355} \\
\bottomrule
\end{tabular}
\end{table*}

\section{Results}
The results examine how target-acquisition exposure affects absolute performance, modality ranking, backbone behaviour, split dependence, robustness, and coverage-specific accuracy.
\subsection{Primary SegFormer Results}
Table~\ref{tab:segformer} shows a protocol-dependent ranking. When the target acquisition is completely held out, RGB is strongest at $0.7317\pm0.0201$ IoU, followed by MS and fusion. When target images are available during development, RGB+MS becomes strongest at $0.7822\pm0.0269$. The conclusion that fusion is best is therefore supported only under target exposure.
\begin{figure*}[t]
\centering
\includegraphics[width=.90\linewidth]{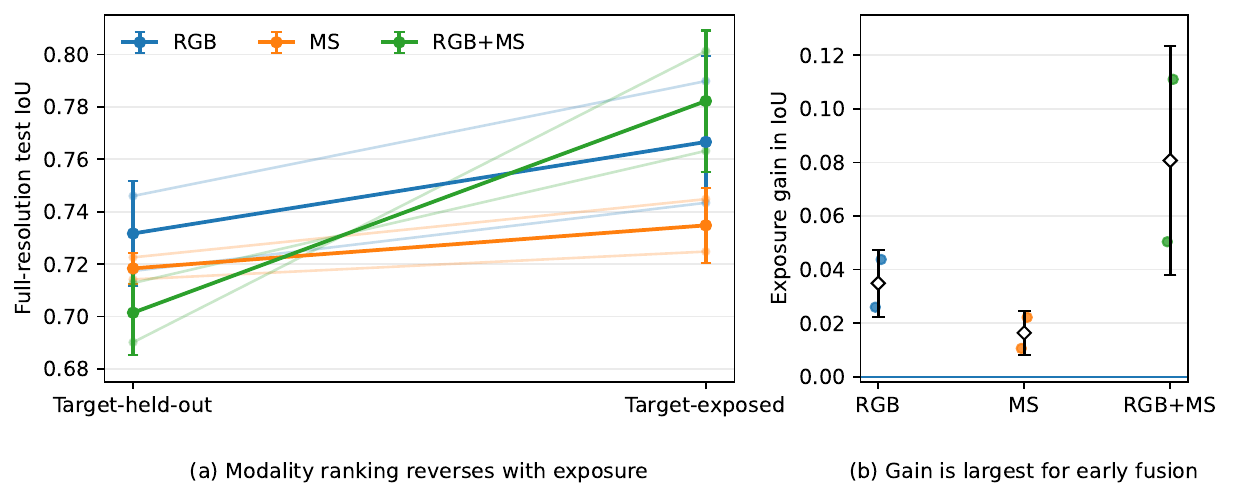}
\caption{Primary SegFormer protocol effect. In panel (a), thick lines show two-seed means and faint lines show the individual fixed-split seeds. In panel (b), colored points show per-seed exposure gains and hollow diamonds show the mean $\pm$ seed standard deviation.}
\label{fig:primary}
\end{figure*}
\begin{table}[ht]
\caption{Fixed-split backbone replication at seed 17. $\Delta$ is exposed minus held-out IoU.}
\label{tab:backbone}
\centering
\small
\setlength{\tabcolsep}{4pt}
\begin{tabular}{@{}llccc@{}}
\toprule
Backbone & Input & Held-out & Exposed & $\Delta$ \\
\midrule
\multirow{3}{*}{SegFormer-B0}
& RGB    & .7174 & .7434 & +.0260 \\
& MS     & .7142 & .7248 & +.0106 \\
& RGB+MS & .7128 & .7632 & +.0504 \\
\midrule
\multirow{3}{*}{U-Net/ResNet18}
& RGB    & \textbf{.7381} & \textbf{.7869} & +.0489 \\
& MS     & .6937 & .7403 & +.0467 \\
& RGB+MS & .7211 & .7374 & +.0164 \\
\bottomrule
\end{tabular}
\end{table}

\subsection{Exposure Gains and Fusion--RGB Interaction}
All SegFormer inputs improve under target exposure in both seeds. Mean gains are $0.0350\pm0.0127$ IoU for RGB, $0.0164\pm0.0083$ for MS, and $0.0808\pm0.0430$ for fusion (Figure~\ref{fig:primary}b). More importantly, Eq.~\eqref{eq:interaction} is positive in both runs: $+0.0244$ with 95\% CI $[0.0164,0.0321]$ for seed 17 and $+0.0673$ $[0.0565,0.0779]$ for seed 43. Under SegFormer, exposure therefore increases fusion's relative advantage over RGB rather than shifting every modality equally. Pairwise modality bootstraps further show that exposed RGB+MS exceeds RGB and MS in both seeds, whereas held-out modality differences are smaller and less consistent.

\subsection{U-Net/ResNet18 Backbone Replication}
Table~\ref{tab:backbone} reports a fixed-split seed-17 replication. Target exposure improves all U-Net inputs, with paired-bootstrap gains of $+0.0489$ $[0.0421,0.0565]$ for RGB, $+0.0467$ $[0.0290,0.0632]$ for MS, and $+0.0164$ $[0.0042,0.0279]$ for fusion. Unlike SegFormer, RGB remains best under both protocols. The fusion--RGB interaction is negative for U-Net ($-0.0325$, 95\% CI $[-0.0456,-0.0206]$) and differs from the SegFormer seed-17 interaction by $+0.0569$ $[0.0442,0.0701]$. Thus, acquisition exposure increases measured performance in both evaluated backbones, but its influence on modality ranking is architecture-dependent.
\begin{figure}[ht]
\centering
\includegraphics[width=.70\linewidth]{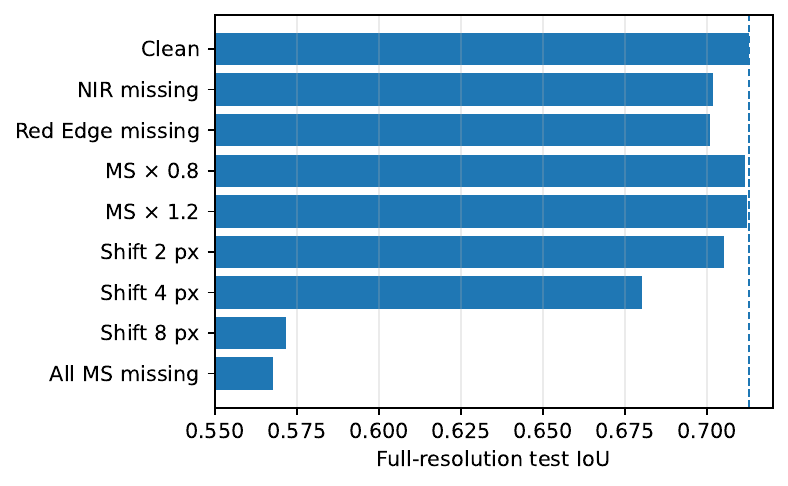}
\caption{Inference-only corruption of the seed-17 held-out SegFormer RGB+MS model. The dashed line marks clean IoU.}
\label{fig:robustness}
\end{figure}

\subsection{Supplied-Split Dependence and Sensor Robustness}
The supplied image-level split assigns every acquisition across training, validation, and testing. For its 148 validation and 148 test images, the median nearest same-date training-frame gap is one. Approximately 98.65\% of validation images and 100\% of test images have a training frame within two frame numbers, and the median nearest training GPS distance is 5.42--5.44~m. Difference hashes do not support a claim of pixel-identical duplication. The defensible conclusion is strong acquisition and near-sequential dependence, under which an image-level split may overstate transfer to a new survey.

Figure~\ref{fig:robustness} summarizes the inference-only corruption analysis of the seed-17 held-out SegFormer RGB+MS model. The results show a clear
asymmetry between radiometric and geometric failure. Removing NIR or Red Edge costs only 0.011--0.012 IoU, and scaling all MS bands by 0.8 or 1.2 changes IoU by at most 0.0014. A four-pixel displacement costs 0.0327; an eight-pixel displacement and complete MS loss cost 0.1411 and 0.1450, respectively. Early fusion is therefore more sensitive to geometric alignment and complete modality failure than to moderate global radiometric scaling.

\subsection{Coverage Dependence and Compute Cost}
Performance increases with annotated weed coverage. Averaged over the 12 primary SegFormer protocol--modality--seed cells, mean image IoU is 0.564 for
the eight low-coverage images ($\leq 0.30$), 0.662 for the 62 medium-coverage images, and 0.817 for the 54 high-coverage images (see Appendix~\ref{sec:coverage}, Fig.~\ref{fig:coverage_iou}, and Table~\ref{tab:coverage}). Because the low-coverage group contains only eight images, this analysis should be interpreted as exploratory and does not provide strong evidence of reliable early-infestation performance. The 12 SegFormer runs required 7.48 Tesla-T4 GPU-hours for training and
0.41 hours for full-resolution validation/test evaluation, whereas the six U-Net runs required 4.21 and 0.20 hours, respectively.
\subsection{Cross-Backbone Interaction Analysis}
The fusion--RGB interaction is positive for SegFormer-B0 but negative for U-Net/ResNet18, even though both backbones benefit from target exposure. This contrast supports an architecture-dependent interpretation rather than a universal advantage for seven-channel fusion. Appendix~\ref{sec:interaction_appendix} provides the interaction visualization.

\section{Discussion}
The discussion interprets the experimental findings in terms of deployment, modality choice, backbone dependence, and the distinction between unseen-acquisition transfer and target-informed adaptation.
\subsection{What the same-test comparison reveals}
The most stable result is that target-acquisition exposure increases measured performance. The effect is positive for every input in both SegFormer seeds and in the U-Net replication. Because the protocols use the same target test images and equal development-set sizes, the gain cannot be explained by an easier test distribution or a larger training set. It instead measures the practical value of having labelled examples from the deployment acquisition. The two protocols therefore answer different deployment questions. The acquisition-mixed setting reflects performance after the model has seen examples from the survey of interest, whereas the held-out setting tests whether it can be carried to a survey that was not represented during development.

The modality ranking, however, does not transfer cleanly from one backbone to the other. SegFormer moves from RGB in the held-out setting to RGB+MS after exposure, but U-Net continues to favour RGB in both cases. The fusion--RGB interaction consequently changes sign across the two backbones. One possible interpretation is that SegFormer's hierarchical attention makes greater use of long-range, pixel-aligned spectral cues once target examples are present, whereas the U-Net model benefits more consistently from visible texture. This remains an interpretation rather than a demonstrated mechanism. What the results do show is that the advantage of spectral fusion cannot be stated independently of the architecture and evaluation protocol.

\subsection{Implications for multimodal deployment}
The corruption tests point to two practical requirements for deployment. First, the fusion model tolerates moderate changes in MS intensity after normalization, yet its accuracy falls steadily as RGB and MS become misregistered. Second, replacing all MS bands with zeros does not make the network behave like the independently trained RGB model; IoU drops by about 0.145. In other words, early concatenation depends on both reliable registration and continued access to the spectral sensor. A deployed system should therefore check alignment and band validity and specify what happens when one modality fails. Training with modality dropout, using later fusion, explicitly aligning features, or maintaining separate unimodal models are reasonable directions for improving this behaviour.

The findings also suggest a conditional deployment strategy. When a small labelled calibration set from the new acquisition is available and registration has been verified, fusion may provide useful gains, particularly for SegFormer. When no labelled target-acquisition examples are available, RGB was the strongest input in both evaluated backbones. This suggests RGB as a reasonable baseline for unseen-acquisition transfer, while avoiding a universal sensor ranking.

\section{Limitations and Reproducibility}
Several limitations remain. First, the target differs from source acquisitions in date, location, weed prevalence, phenotype, illumination, and altitude; the observed shift cannot be attributed to one variable. Second, the target test set is weed-dense and contains only eight images below 0.30 coverage, limiting conclusions about sparse early infestation. Third, SegFormer is repeated with two model seeds, whereas U-Net is a one-seed replication. Additional architectures, seeds, and independent acquisition domains are needed before generalizing the interaction pattern. Fourth, the study evaluates early concatenation only; more sophisticated fusion may respond differently to exposure and misalignment. Finally, polygon annotations can retain spatially structured uncertainty even after expert review~\cite{nguyen2025dataset}.

The experimental safeguards are nevertheless strong for a compact benchmark study. The split is fixed before cropping; normalization statistics are estimated only from training data; thresholds are selected only on validation images; every protocol and modality comparison uses the same 124 test identifiers; and per-image confusion counts support paired resampling. Checkpoints, split manifests, threshold curves, software versions, and table-generation artifacts are retained. The complete 18-run experiment required approximately 11.69 T4 GPU-hours for training and 0.61 hours for full-resolution evaluation, making independent replication feasible.

\section{Conclusion}
Acquisition exposure consistently increases measured UAV weedy-rice segmentation performance across the two evaluated backbones, but the apparent value of multimodal input depends on the backbone. SegFormer exhibits a protocol-dependent ranking reversal, with RGB strongest under complete acquisition holdout and RGB+MS strongest after target exposure. U-Net confirms positive exposure gains but retains RGB as the best input under both protocols. Thus, acquisition exposure is a general evaluation concern in these experiments, whereas fusion-specific conclusions are architecture-dependent. Together with the supplied-split audit, coverage analysis, and misregistration tests, these findings support acquisition-aware same-test evaluation as a necessary complement to ordinary image-level splitting in multimodal UAV benchmarks.

\bibliographystyle{plainnat}
\bibliography{refs}
\appendix
\section{Cross-Backbone Interaction Visualization}
\label{sec:interaction_appendix}
\begin{figure}[ht]
\centering
\includegraphics[width=.90\linewidth]{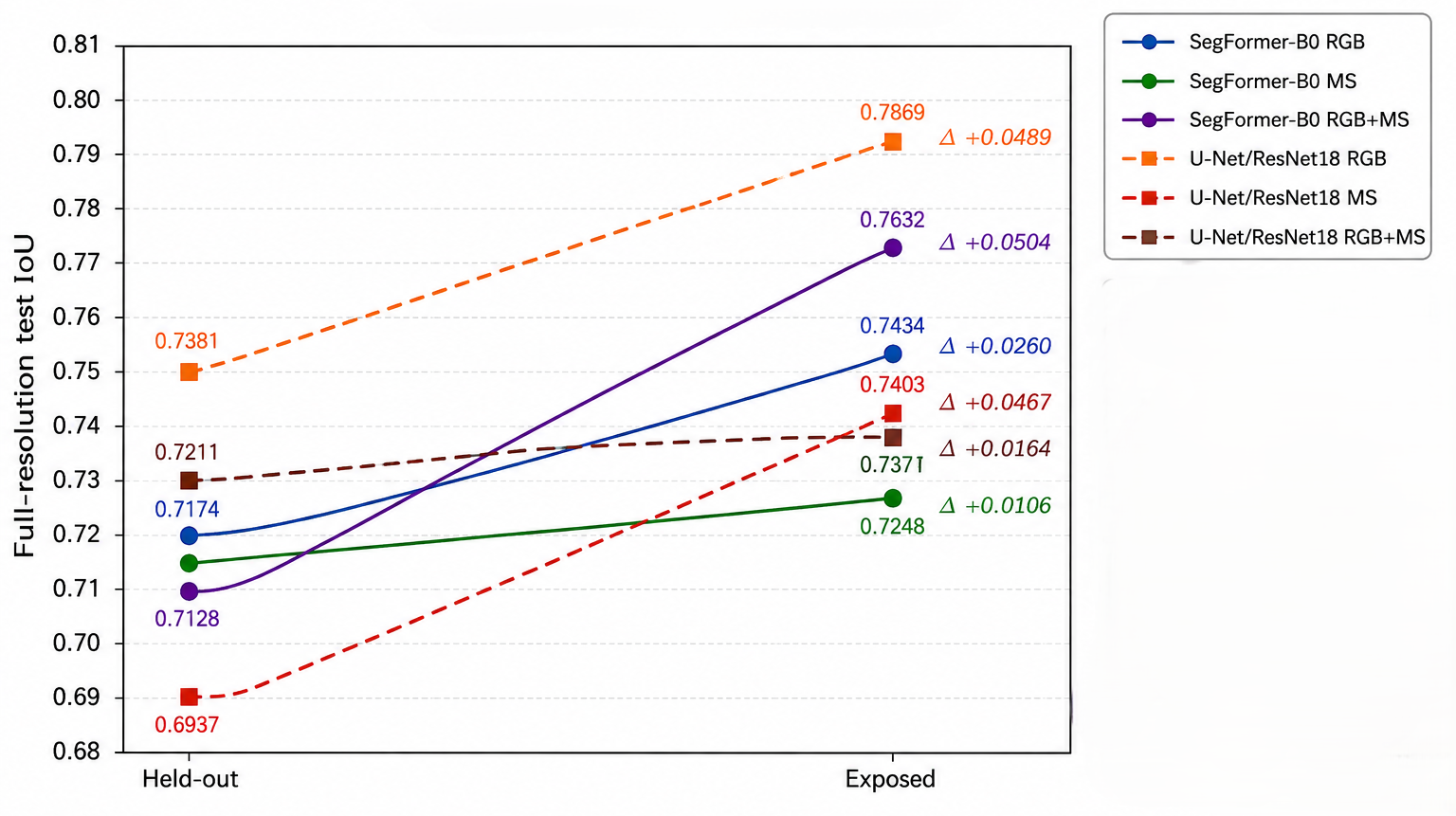}
\caption{Backbone $\times$ acquisition-exposure comparison at seed 17 on the fixed target test set. The largest exposure gain occurs for SegFormer-B0 fusion and for U-Net/ResNet18 RGB.}
\label{fig:backbone_exposure}
\end{figure}
\section{Dataset Analysis}
\label{sec:dataset}
\begin{figure*}[ht]
\centering
\includegraphics[width=.90\linewidth]{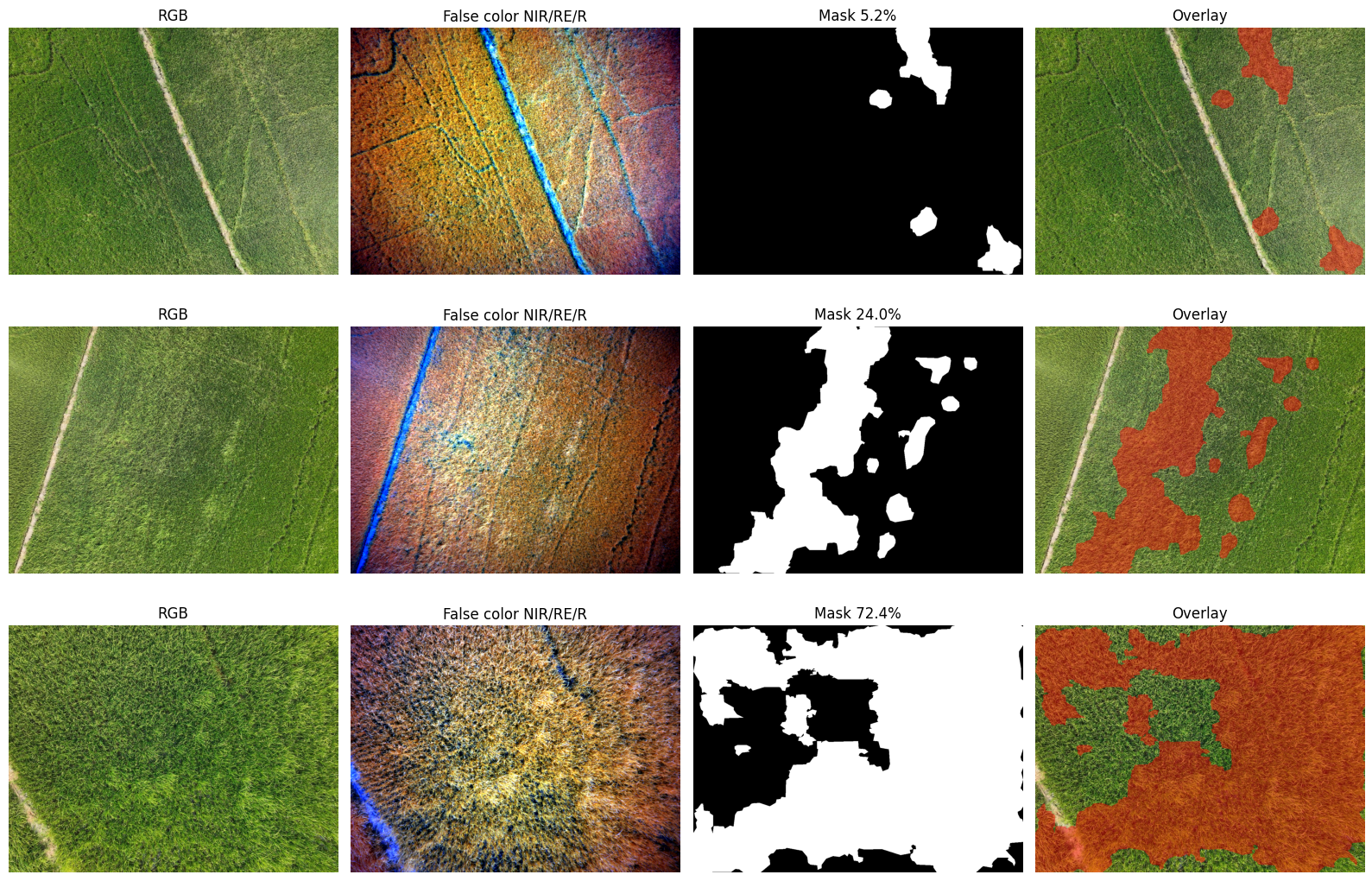}
\caption{Representative low-, medium-, and high-coverage samples with annotated weed coverage of 5.2\%, 24.0\%, and 72.4\%, respectively. Columns show RGB, false-color NIR/Red Edge/Red, ground-truth mask, and mask overlay. The visual appearance and annotated prevalence vary
substantially across the benchmark.}
\label{fig:examples}
\end{figure*}

To further characterize the visual diversity of the benchmark, we examine representative samples spanning low, medium, and high weed coverage. As illustrated in Fig.~\ref{fig:examples}, the dataset contains substantial
variation in weed prevalence and visual appearance across acquisitions. The RGB imagery, false-color NIR/Red Edge/Red composites, and corresponding ground-truth masks reveal differences in spatial distribution, vegetation
density, and background composition. These variations provide a useful basis for evaluating model robustness beyond aggregate segmentation performance.
\begin{figure*}[ht]
    \centering
    \includegraphics[width=.95\linewidth]{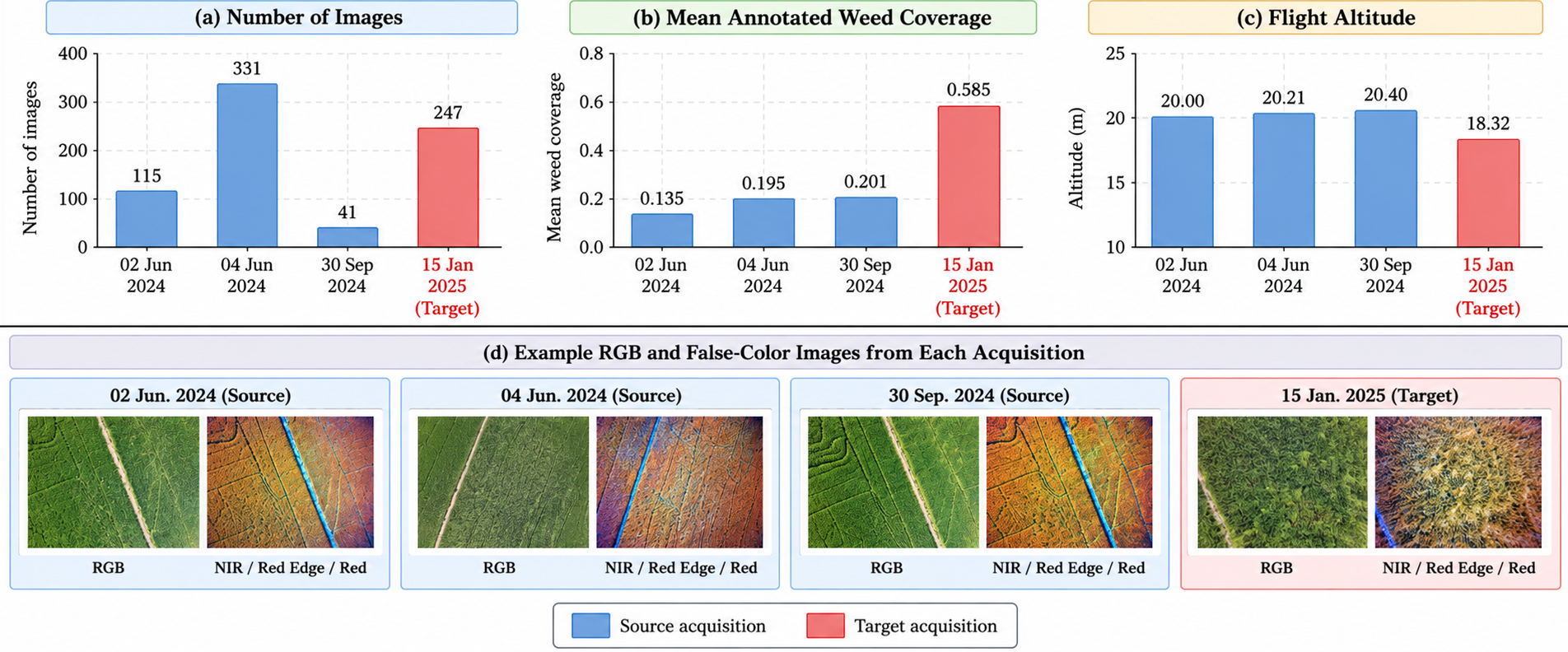}
    \caption{Acquisition-level characteristics of WeedyRice-RGBMS-DB.
    The target acquisition differs from the source acquisitions in sample
    count, annotated weed coverage, and flight altitude, illustrating the
    acquisition-domain shift considered in this study.}
    \label{fig:acquisition_shift}
\end{figure*}
\begin{table}[t]
\caption{Acquisition inventory and supplied-split dependence.}
\label{tab:inventory}
\centering
\small
\setlength{\tabcolsep}{2pt}
\begin{tabular}{@{}lrrr@{}}
\toprule
Acquisition & Images & Mean cov. & Altitude (m) \\
\midrule
02 Jun. 2024 & 115 & 0.135 & 20.00 \\
04 Jun. 2024 & 331 & 0.195 & 20.21 \\
30 Sep. 2024 & 41  & 0.201 & 20.40 \\
15 Jan. 2025 (target) & 247 & 0.585 & 18.32 \\
\midrule
\multicolumn{4}{@{}l@{}}{\textit{Supplied validation/test audit}} \\
Median same-date frame gap
    & \multicolumn{3}{c@{}}{1 frame} \\
Within two training frames
    & \multicolumn{3}{c@{}}{98.65\% / 100\%} \\
Median nearest train GPS
    & \multicolumn{3}{c@{}}{5.42 / 5.44 m} \\
\bottomrule
\end{tabular}
\end{table}

In addition to visual variation, the dataset exhibits substantial differences across acquisition dates. Figure~\ref{fig:acquisition_shift} summarizes
differences in sample count, annotated weed coverage, and flight altitude between the source and target acquisitions. Table~\ref{tab:inventory} provides the corresponding numerical values and supplied-split dependence statistics.
\section{Additional Coverage-Dependent Results}
\label{sec:coverage}
\begin{figure}[ht]
    \centering
    \includegraphics[width=.75\linewidth]{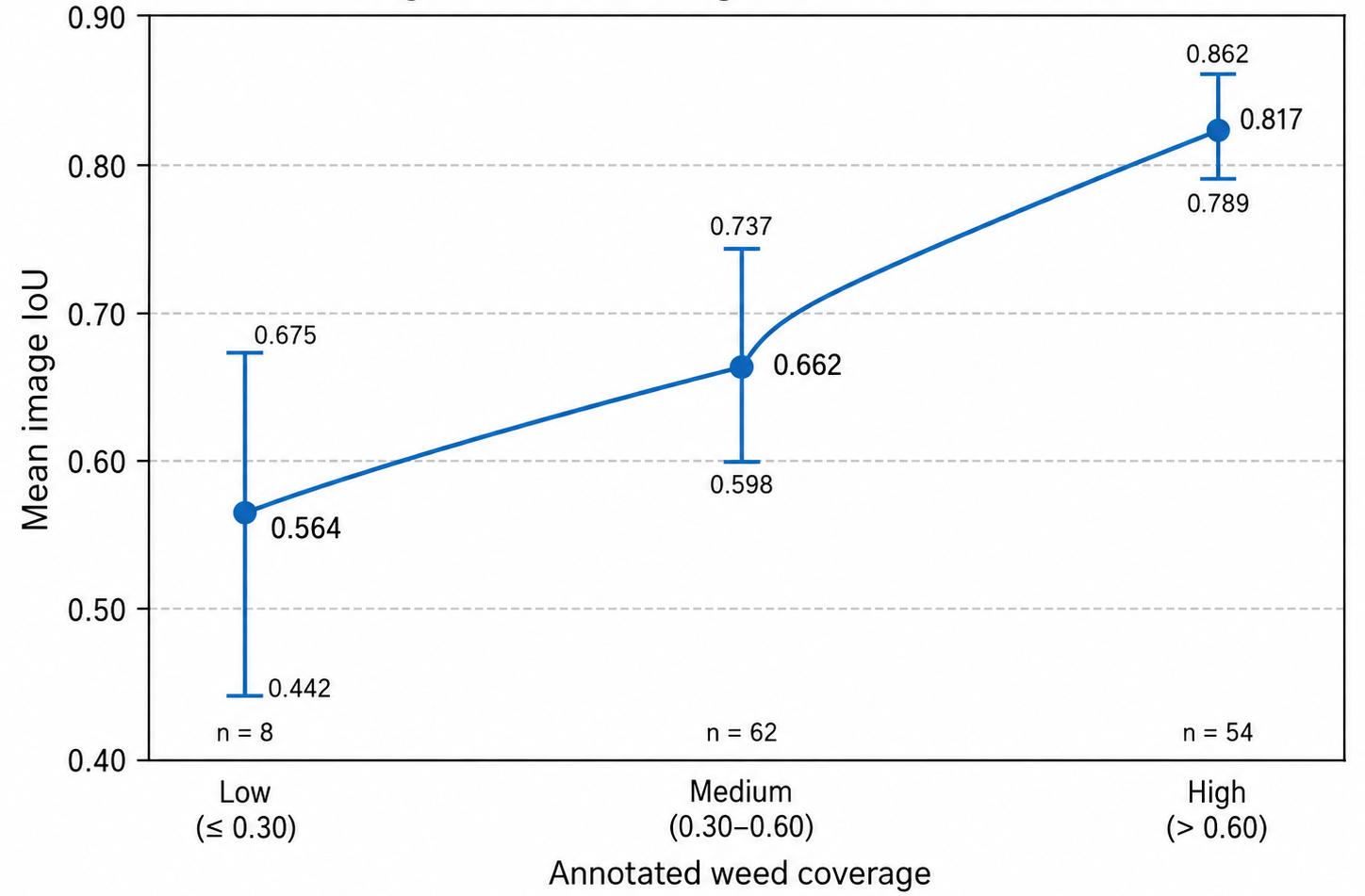}
    \caption{Coverage-stratified segmentation performance across low,
    medium, and high annotated weed-coverage groups. Points show mean image
    IoU, while whiskers indicate the reported cell range across the 12 primary
    SegFormer protocol--modality--seed cells.}
    \label{fig:coverage_iou}
\end{figure}

This appendix provides the detailed coverage-stratified results supporting the
analysis presented in the main text. Figure~\ref{fig:coverage_iou} visualizes
the performance trend, while Table~\ref{tab:coverage} reports the exact values.
\begin{table}[!htbp]
\caption{Coverage-stratified image IoU for the primary SegFormer cells.}
\label{tab:coverage}
\centering
\small
\begin{tabular}{lrrc}
\toprule
Weed coverage & Images & Mean IoU & Cell range \\
\midrule
$\leq 0.30$ & 8  & 0.564 & 0.442--0.675 \\
0.30--0.60  & 62 & 0.662 & 0.598--0.737 \\
$> 0.60$    & 54 & 0.817 & 0.789--0.862 \\
\bottomrule
\end{tabular}
\end{table}
\end{document}